\documentclass[pdflatex,sn-mathphys-num]{sn-jnl}
\usepackage{graphicx}%
\usepackage{multirow}%
\usepackage{amsmath,amssymb,amsfonts}%
\usepackage{amsthm}%
\usepackage{mathrsfs}%
\usepackage[title]{appendix}%
\usepackage{xcolor}%
\usepackage{textcomp}%
\usepackage{booktabs}
\usepackage{manyfoot}%
\usepackage{booktabs}%
\usepackage{algorithm}%
\usepackage{algorithmicx}%
\usepackage{algpseudocode}%
\usepackage{listings}%
\usepackage{ulem}
\usepackage{xcolor}
\definecolor{darkyellow}{rgb}{0.7, 0.7, 0} 
\definecolor{vividblue}{rgb}{0, 0.2, 0.8}

\theoremstyle{thmstyleone}%
\theoremstyle{thmstyletwo}%
\theoremstyle{thmstylethree}%
\begin{document}

\title[Article Title]{From Pre- to Intra-operative MRI: Predicting Brain Shift in Temporal Lobe Resection for Epilepsy Surgery}

\author[1]{\fnm{Jingjing} \sur{Peng}}
\author[4]{\fnm{Giorgio} \sur{Fiore}}
\author[1]{\fnm{Yang} \sur{Liu}}
\author[1]{\fnm{Ksenia} \sur{Ellum}}
\author[2]{\fnm{Debayan} \sur{Daspupta}}
\author[3]{\fnm{Keyoumars} \sur{Ashkan}}
\author[2]{\fnm{Andrew} \sur{McEvoy}}
\author[2]{\fnm{Anna} \sur{Miserocchi}}
\author[1]{\fnm{Sebastien} \sur{Ourselin}}
\author[2]{\fnm{John} \sur{Duncan}}
\author*[1]{\fnm{Alejandro} \sur{Granados}}\email{alejandro.granados@kcl.ac.uk}

\affil*[1]{\orgdiv{Surgical and Interventional Engineering}, \orgname{King's College London}, \country{UK}}
\affil[2]{\orgdiv{National Hospital for Neurology and Neurosurgery}, \orgname{NHS}, \country{UK}}
\affil[3]{\orgdiv{Neurosurgery, King's College Hospital}, \orgname{NHS}, \country{UK}}
\affil[4]{\orgdiv{Fondazione IRCCS Ca’ Granda}, \orgname{Ospedale Maggiore di Milano}, \country{Italy}}

\abstract{

\textbf{Introduction:} In neurosurgery, image-guided Neurosurgery Systems (IGNS) highly rely on preoperative brain magnetic resonance images (MRI) to assist surgeons in locating surgical targets and determining surgical paths. However, brain shift invalidates
preoperative MRI after dural opening. Updated intraoperative brain MRI with brain shift compensation is crucial for enhancing the precision of neuronavigation systems and ensuring the optimal outcome of surgical interventions. 
\textbf{Methodology:} We propose \textbf{NeuralShift}, a U-Net-based model that predicts brain shift entirely from pre-operative MRI for patients undergoing temporal lobe resection. We evaluated our results using Target Registration Errors (TREs) computed on anatomical landmarks located on the resection side and along the midline, and DICE scores  comparing predicted intraoperative masks with masks derived
from intraoperative MRI. \textbf{Results:} Our experimental results show that our model can predict the global deformation of the brain (DICE of 0.97) with accurate local displacements (achieve landmark TRE as low as 1.12 mm),  compensating for large brain shifts during temporal lobe removal neurosurgery.
\textbf{Conclusion:} Our proposed model is capable of predicting the global deformation of the brain during temporal lobe resection using only preoperative images, providing potential opportunities to the surgical team to increase safety and efficiency of neurosurgery and better outcomes to patients. Our contributions will be publicly available after acceptance in \href{https://github.com/SurgicalDataScienceKCL/NeuralShift}{https://github.com/SurgicalDataScienceKCL/NeuralShift}.}

\keywords{Brain shift prediction, Displacement field, Intra-operative MRI, Temporal lobe resection}



\maketitle

\section{Introduction}\label{sec1}

Neuronavigation is routinely used during neurosurgical interventions to align preoperative images with the patient's brain anatomy~\cite{thomas2015image}, allowing surgeons to execute their surgical plans with greater precision. However, the occurrence of brain shift after dura opening undermines this alignment~\cite{gerard2017brain}. Brain shift is multifactorial and is caused by the drainage of cerebrospinal fluid, gravity, pharmacological interventions, changes in cerebral blood volume, and manipulation and resection of tissue, among other factors~\cite{gerard2021brain,gerard2017brain}. Consequently, brain shift not only results in a discrepancy between preoperative magnetic resonance imaging (MRI) and the actual shape of the patient's brain during surgery~\cite{iversen2018automatic,khoshnevisan2012neuronavigation}, but also increases the likelihood of surgical complications due to a reduction of the accuracy of the neuronavigation system. Therefore, there is an urgent need for models that accurately predict brain shift intraoperatively, allowing surgeons to adjust surgical plans accordingly while improving patient outcomes.

While intraoperative MRI (iMRI) allows to comprehensively capture brain shift at a given time during surgery, it poses challenges for its successful deployment into standard practice. iMRI is complex, time-consuming, costly, disruptive, and not commonly available in operating rooms. 
In contrast, intraoperative ultrasound (iUS) is affordable, easier to integrate to a surgical workflow, and faster to acquire, although with the caveat that it provides lower-contrast images, is highly operator-dependent, and is difficult to quantify different lesions' echogenicity objectively~\cite{dixon2022iUS}.

Biomechanical models have traditionally been investigated for brain shift compensation~\cite{ji2009model}. These approaches typically require the integration of registration techniques and intraoperative imaging modalities to update preoperative brain images~\cite{bayer2017intraoperative,correa2017neurosurgery}. 
Morin et al.~\cite{morin2017brain} proposed the use of biomechanical simulations using preoperative data constrained by iUS to account for brain deformation observed during surgery. 
Other approaches either integrate a laser range scanner or a stereo camera~\cite{fan2016intraoperative} with a linear elastic finite element model (FEM) to predict and correct brain shift in real-time. While these approaches eliminate the need for contact and physical connections~\cite{zhuang2011sparse}, they are limited to capturing the deformation of the brain on the surface only, resulting in reduced effectiveness when mapping the deformation of deeper brain structures.

Furthermore, the finite element method employed to deduce global deformations from surface deformations hinges on assumptions regarding the biomechanical properties of human brain, (hyper)elastic models that are only tested on brain specimens, and on simplified boundary conditions~\cite{granados2021hyperelastic}. Such approximations misrepresent the true physiological state of the brain in patient-specific cases~\cite{yang2023finite}. While biomechanical models are promising tools, they rarely run in real-time and are not always clinically compatible~\cite{miga2016clinical}, limiting their utility during surgery.

These limitations have led to the development of approaches that utilise deep learning to learn soft tissue deformation entirely from data.

Pfeiffer et al. presented a fully convolutional neural network (CNN) to estimate organ deformation from a partial view of the organ's surface using simulation-based synthetic data~\cite{pfeiffer2019learning}. This approach generalises well to new patients without the need for retraining, while operating in real-time, and offering comparable accuracy to traditional registration methods. However, the model is limited since the coarse discretisation of the mesh leads to an imprecise representation of the organ's surface and the model operates under the assumption that boundary conditions and surface correspondences can be predetermined.
Liu et al. introduced a real-time artificial neural network that predicts the deformation of regions of interest (ROI) in soft tissues facilitated by sparsely registered fiducial markers, although highly dependent on the quality and quantity of fiducial markers used during surgery~\cite{liu2020real}. 
Lampen et al. proposed a biomechanics-informed PointNet++ architecture that takes point cloud data and explicit boundary types as inputs to predict soft-tissue deformation as a surrogate of FEM, significantly reducing simulation time~\cite{lampen2022deep}. 

However, to the best of our knowledge, only one study has proposed a deep learning model for brain tissue deformation. Shimamoto et al.~\cite{shimamoto2023precise} introduced W-Net, a CNN that compensates for brain shift after dural opening. A significant reduction in target registration error (TRE), at both the tumor center and the maximum shift position, shows the potential of CNNs in correcting brain shift. However, this compensation is only targeted before any resection, rather than during or after the removal of the tumor.
With the exception of RESECT~\cite{xiao2017resect}, and more recently ReMIND~\cite{juvekar2024dataset}, the largest publicly available dataset of surgically treated brain tumours, the limited availability of brain datasets that are able to capture the dynamics of brain shift curtail the opportunities to investigate the predicting capabilities of deep learning in compensating for brain shift.

Recent works have begun to integrate biomechanical simulations to predict intraoperative changes in brain tissue from pre-operative images, sometimes within clinically feasible timeframes (e.g., under three minutes) ~\cite{safdar2023slicercbm}. Although these methods remain focused on registration or deformation estimation, many adopt a U-Net-like multi-scale architecture for learning mappings from pre-op to intra-op states. Similar approaches for real-time MRI reconstruction also rely on U-Net-based networks to handle undersampled k-space data 
~\cite{ottesen2025deep}. While our current study does not address reconstruction problem, it reaffirms that U-Net remains a strong template for tasks requiring robust feature extraction under challenging constraints.

To address these challenges, we propose \textbf{NeuralShift} with the goal of predicting displacement vector fields that map the deformation of the brain between preoperative MRI (pMRI) images and iMRI images in the context of temporal lobe resection for epilepsy surgery.
To achieve this goal, we argue that predicting brain shift can be split into two tasks: a) an investigation of the predicting capabilities of a model that learns brain shift based only on preoperative data as inputs, and b) an extension of this model by considering intraoperative observations. In this paper, we investigate the former. 
Our main contributions are: 1) a preoperative-to-intraoperative registration pipeline that accounts for the resection cavity, while generating deformation fields that quantitatively capture brain shift, 2) a U-Net-based model optimised to learn global and local deformations of the brain on patient-specific cases, and 3) a rigorous validation supported by ablation studies and landmarks-based target registration errors.

\section{Methods}\label{sec2}

\subsection{Temporal Lobe Resection for Epilepsy Surgery Dataset}
We curated a dataset of 98 paired preoperative (pMRI) and intraoperative (iMRI)
T1-weighted MRI scans from epilepsy patients undergoing left or right temporal
lobe resection at the National Hospital for Neurology and Neurosurgery (London, UK). All data were retrospectively evaluated. As a retrospective study on anonymised clinical data, individual patient consent was not required (Health Research Authority 22/SC/0016).
Intraoperative MRI
was acquired after tissue resection and prior to surgical closure, capturing
substantial brain deformation associated with temporal lobe surgery. Temporal
lobe resection represents a clinically relevant setting for studying brain
shift due to the presence of consistent surgical workflows and pronounced
deformation patterns.

All experiments were conducted using 9-fold cross-validation. The dataset was
randomly partitioned into nine folds, with each fold used once for evaluation
while the remaining folds were used for training. This strategy ensures that
all subjects contribute to both training and evaluation, providing a robust
estimate of model performance given the limited cohort size. 

Resection laterality information was retrospectively derived and used to
construct a brain hemisphere indicator encoding whether the left or right temporal
lobe was resected. This indicator provides coarse laterality information without
incorporating details about the exact resection geometry or extent.

\subsection{Pre-to-Intraoperative Registration Pipeline}
\begin{figure}
\centering
\includegraphics[width=\linewidth]{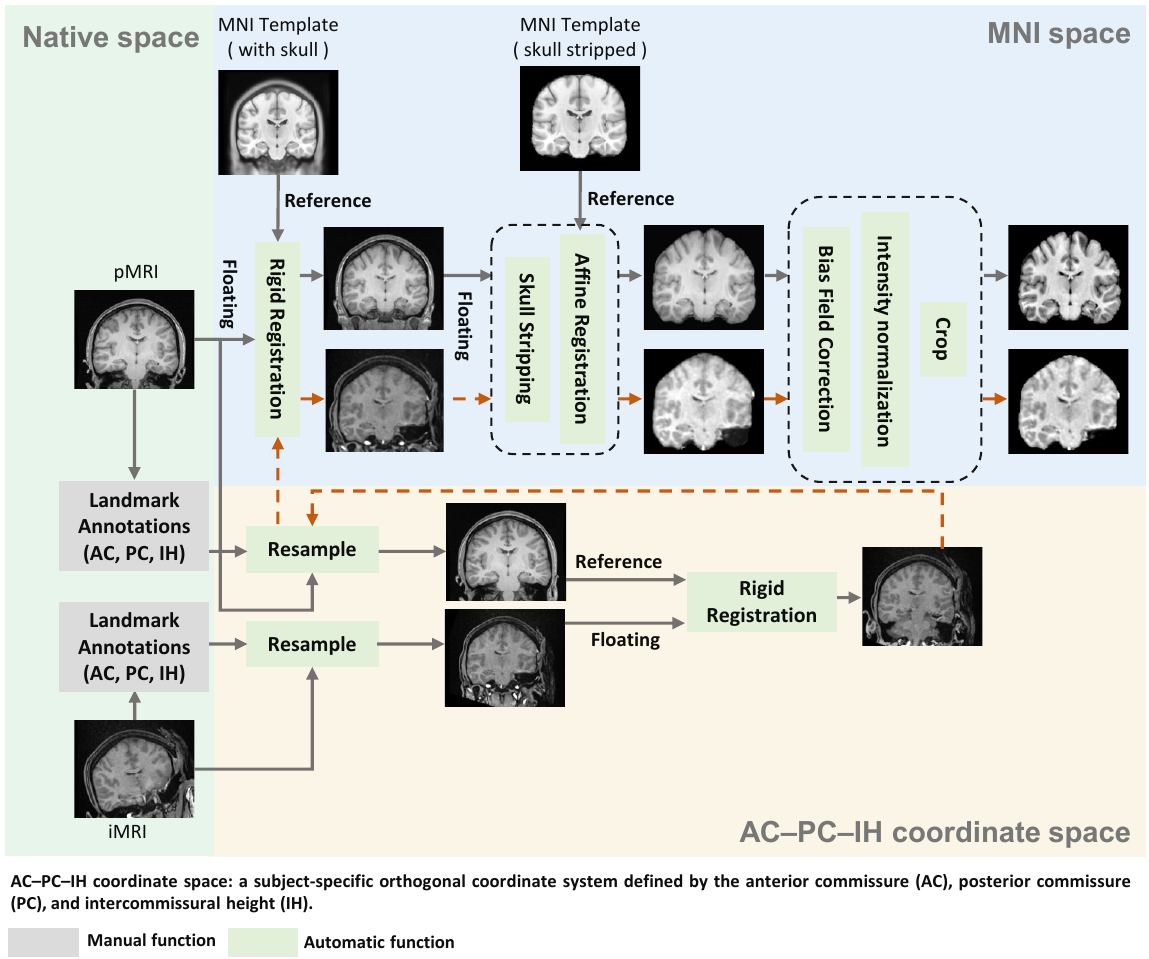}
\caption{Preoperative-to-intraoperative MRI preprocessing pipeline. Preoperative MRI (pMRI) is rigidly aligned to the MNI template, skull stripped, and affinely registered. Intraoperative MRI (iMRI), containing the post-resection cavity, is first reoriented using a subject-specific AC--PC--IH coordinate system defined by manually annotated landmarks, and then rigidly registered to the corresponding pMRI in native space. The pMRI-to-MNI transformations are subsequently propagated to iMRI so that both modalities share a common MNI space. Finally, intensity normalisation, bias-field correction, and cropping are applied to produce standardised network inputs.}

\label{fig:pipeline}
\centering
\end{figure}

Compared to preoperative MRI (pMRI), intraoperative MRI (iMRI) acquired after temporal lobe resection contains a surgical cavity and intensity heterogeneity,
which complicates the direct application of standard registration tools for normalisation to a common stereotaxic space. In this work, we use the Montreal Neurological Institute (MNI) template, a population-averaged brain atlas that provides a widely adopted anatomical reference space for spatial normalisation and inter-subject comparison in neuroimaging studies.

To robustly align both pMRI and iMRI images into MNI space, we designed the bespoke preprocessing pipeline illustrated in Fig.~\ref{fig:pipeline}. We first
manually identified three anatomical landmarks---the anterior commissure (AC), posterior commissure (PC), and intercommissural height (IH)---on both pMRI and iMRI
images. These landmarks define a subject-specific orthogonal AC--PC--IH coordinate system, which is used to resample the images and standardise their orientation prior to further registration. This landmark-based reorientation mitigates inter-subject variability and improves the robustness of subsequent rigid
registration, particularly in the presence of resection cavities in iMRI.

Following this initial alignment, iMRI images are rigidly registered to their corresponding pMRI images in native space. The pMRI images are then rigidly
registered to the MNI template with skull, followed by skull stripping and affine registration to the skull-stripped MNI template using NiftyReg~\cite{modat2014global}. The transformation matrices estimated from the pMRI-to-MNI registration are subsequently propagated to the iMRI images, ensuring that both modalities undergo identical spatial transformations and are ultimately expressed in the same MNI space. Finally, intensity normalisation, bias-field correction, and cropping are applied to obtain standardised inputs suitable for learning-based deformation modelling.

\subsection{Model Architecture and Loss Functions}
\begin{figure}
\centering
\includegraphics[width=0.95\linewidth]{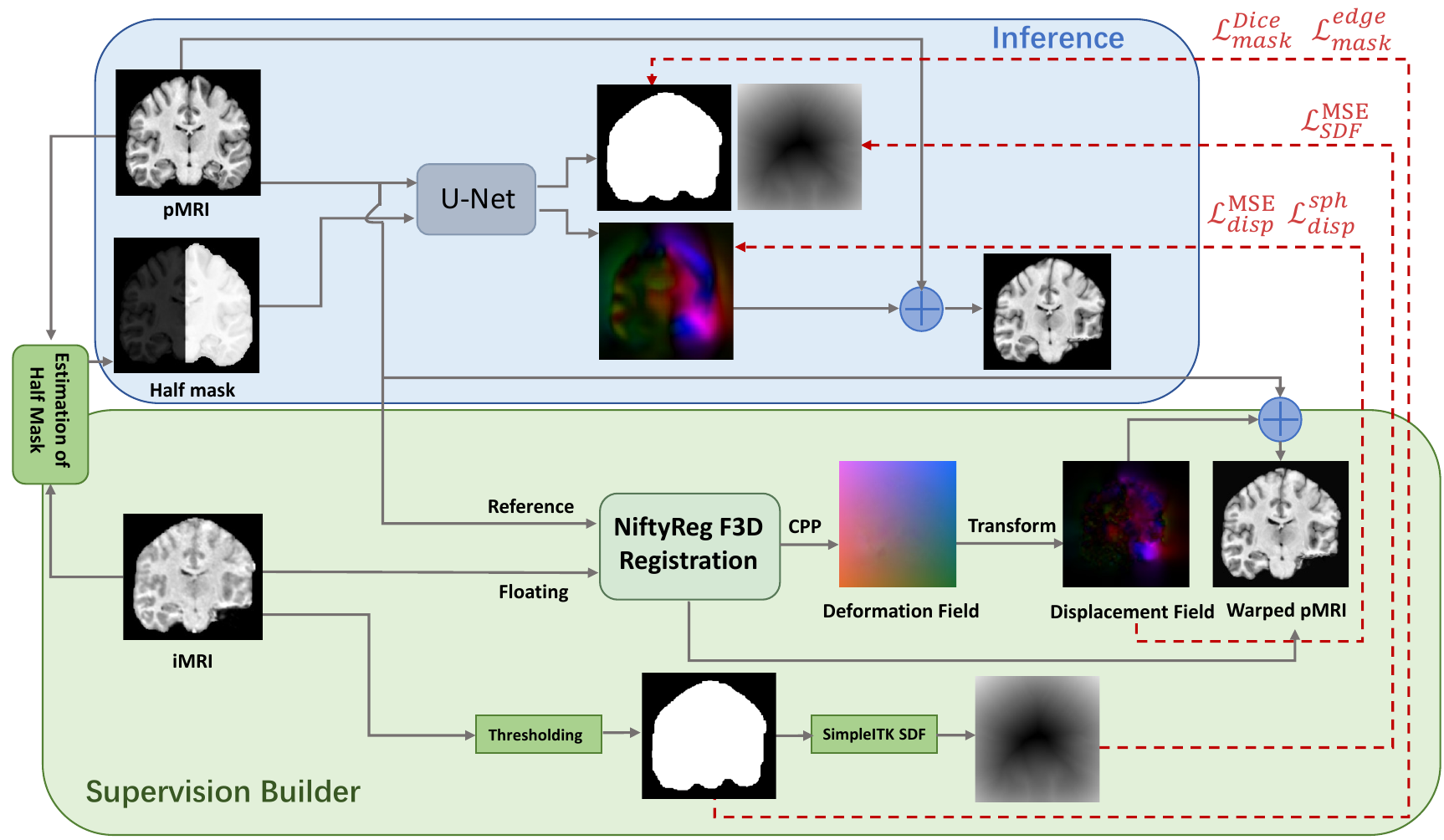}
\caption{Overview of the proposed brain shift prediction framework. The inference module takes as input the preoperative MRI (pMRI) and a hemisphere indicator (``half mask'') encoding resection laterality, and predicts (i) a dense displacement field mapping pMRI to iMRI space, (ii) the intraoperative brain mask, and (iii) its signed distance function (SDF) using a U-Net. The supervision builder constructs training targets via non-rigid registration (NiftyReg F3D) between pMRI and iMRI, yielding a dense displacement field used to warp pMRI. The intraoperative mask is obtained by thresholding, and the corresponding SDF is computed from the mask. Training employs separate loss terms for displacement field regression, mask prediction, and SDF supervision.}

\label{fig:Overview architecture}
\centering
\end{figure}

Our framework consists of an inference module and a supervision builder (Fig.~\ref{fig:Overview architecture}). Rather than synthesising intraoperative MRI (iMRI) intensities, we formulate brain-shift prediction as supervised regression of a dense displacement field mapping preoperative MRI (pMRI) to iMRI.

As voxel-wise physical ground truth for brain deformation is unavailable, we adopt a registration-based surrogate supervision strategy. Specifically, pMRI is registered to iMRI using the Fast Free-Form Deformation (F3D) algorithm implemented in NiftyReg~\cite{modat2010fast}, yielding a free-form deformation model parameterised by cubic B-spline control points stored in \texttt{cpp} format. This model implicitly defines a deformation field \( y(x) \), from which the displacement field \( u(x)=y(x)-x \) is derived and used as the regression target. Displacement vectors are expressed in physical units (millimetres) according to image spacing. We emphasise that this target represents a consistent surrogate rather than exact biomechanical ground truth.

The inference network is based on a U-Net~\cite{ronneberger2015u}. It takes as input the pMRI and a hemisphere indicator (``half mask''), defined as a binary mask encoding the planned resection laterality (left or right hemisphere) without specifying resection extent. This design is consistent with a preoperative planning scenario. The network predicts three outputs: the displacement field, the intraoperative brain mask, and the corresponding signed distance function (SDF).

While voxel-wise displacement regression captures local deformation, it provides limited explicit constraints on global brain geometry, particularly in the presence of tissue resection and surgical cavities. To complement local deformation supervision with global shape information, we additionally incorporate supervision on the intraoperative brain mask and its SDF representation. The SDF provides a continuous, dense encoding of the signed distance to the brain boundary, offering stronger geometric constraints than binary mask supervision alone.

Training is performed using a weighted multi-task objective combining displacement, mask, and SDF supervision. Displacement accuracy is enforced using a Cartesian mean squared error (MSE) loss,
\begin{equation}
\mathcal{L}_{\text{disp}}^{\text{MSE}} = \frac{1}{N} \sum_{i=1}^{N}
(\mathbf{v}_{pi} - \mathbf{v}_{gi})^2 ,
\end{equation}
together with an auxiliary spherical-coordinate loss that penalises angular and magnitude discrepancies. Each 3D displacement vector is transformed from Cartesian coordinates into spherical coordinates, parameterised by an elevation angle \(\theta\), an azimuthal angle \(\phi\), and a magnitude. The corresponding loss terms are defined as
\begin{equation}
\mathcal{L}_{\text{disp}}^{\theta} =
\frac{1}{N} \sum_{i=1}^{N}
\big(\theta(\mathbf{v}_{pi}) - \theta(\mathbf{v}_{gi})\big)^2 ,
\end{equation}
\begin{equation}
\mathcal{L}_{\text{disp}}^{\phi} =
\frac{1}{N} \sum_{i=1}^{N}
\min\!\left(
\big(\phi(\mathbf{v}_{pi}) - \phi(\mathbf{v}_{gi})\big)^2 ,
\big(2\pi - \lvert \phi(\mathbf{v}_{pi}) - \phi(\mathbf{v}_{gi}) \rvert\big)^2
\right),
\end{equation}
\begin{equation}
\mathcal{L}_{\text{disp}}^{\text{mag}} =
\frac{1}{N} \sum_{i=1}^{N}
\left(
\lVert \mathbf{v}_{pi} \rVert_2 -
\lVert \mathbf{v}_{gi} \rVert_2
\right)^2 .
\end{equation}
The spherical-coordinate loss is defined as
\begin{equation}
\mathcal{L}_{\text{disp}}^{\text{sph}} =
\mathcal{L}_{\text{disp}}^{\theta} +
\mathcal{L}_{\text{disp}}^{\phi} +
\mathcal{L}_{\text{disp}}^{\text{mag}} .
\end{equation}
Mask prediction is supervised using a Dice loss
\begin{equation}
\mathcal{L}_{\text{mask}}^{\text{Dice}} =
1 - \frac{2 \sum_{i=1}^{N} \hat{M}^i_{\text{intra}} M^i_{\text{intra}}}
{\sum_{i=1}^{N} (\hat{M}^i_{\text{intra}})^2 + \sum_{i=1}^{N} (M^i_{\text{intra}})^2} ,
\end{equation}
together with an edge loss
\begin{equation}
\mathcal{L}_{\text{mask}}^{\text{edge}} =
\frac{1}{N} \sum_{i=1}^{N}
\left|
\text{dilate}(\hat{M}^i_{\text{intra}}, k) -
\text{dilate}(M^i_{\text{intra}}, k)
\right|.
\end{equation}
For SDF prediction, an MSE loss is used:
\begin{equation}
\mathcal{L}_{\text{SDF}}^{\text{MSE}} =
\frac{1}{N} \sum_{i=1}^{N}
(\hat{\mathbf{s}}^i - \mathbf{s}^i)^2 .
\end{equation}

The overall objective is a weighted sum of the individual loss terms:
\begin{equation}
\mathcal{L} =
\alpha \left(
\mathcal{L}_{\text{disp}}^{\text{MSE}} +
\mathcal{L}_{\text{disp}}^{\text{sph}}
\right)
+ \gamma \, \mathcal{L}_{\text{SDF}}^{\text{MSE}}
+ \beta \left(
\mathcal{L}_{\text{mask}}^{\text{Dice}} +
\mathcal{L}_{\text{mask}}^{\text{edge}}
\right),
\end{equation}
where $\alpha$, $\beta$, and $\gamma$ balance the contributions of the displacement, mask, and SDF losses to account for differences in numerical scale.

\subsection{Experimental Design}

We evaluated the proposed method using two complementary metrics: the Dice Similarity Coefficient (Dice) and the Target Registration Error (TRE). Dice measures the overlap between the predicted and ground-truth intraoperative brain masks, with a value of 1 indicating perfect agreement. In contrast, TRE quantifies local geometric accuracy by measuring the distance between corresponding anatomical landmarks after deformation, with lower values indicating better alignment.

For landmark-based evaluation, a total of ten anatomical landmarks were manually identified on the preoperative MRI (pMRI) in native space. These landmarks were selected to represent clinically relevant anatomical structures that are commonly referenced during neurosurgical planning and craniotomy. Specifically, the landmarks include points located at the frontal operculum ($P_{1L}$, $P_{1R}$), the junction between the posterior insula and temporal opercula ($P_{2L}$, $P_{2R}$), the aqueduct of Sylvius ($P_3$), the superior frontal region ($P_{4L}$, $P_{4R}$), the corpus callosum ($P_5$), and the optic tract ($P_{6L}$, $P_{6R}$). Here, the subscripts $L$ and $R$ denote anatomically symmetric landmark locations in the left and right cerebral hemispheres, respectively. All landmarks were annotated using the 3D Slicer software by trained annotators under the guidance of experienced neurosurgeons. For TRE evaluation, we focus on landmarks located on the resection side together with midline landmarks. This choice reflects the clinical characteristics of temporal lobe resection for epilepsy surgery, where larger deformations are typically observed in the vicinity of the surgical corridor, while displacement
on the contralateral side is comparatively small. Moreover, neurosurgical decision-making primarily depends on accurate guidance near the resection region and adjacent critical structures.

To compute TRE, the predicted displacement field was applied to each preoperative landmark, yielding its predicted intraoperative position $\hat{P}^i_{\text{intra}}$. The corresponding ground-truth intraoperative positions $P^i_{\text{intra}}$ were obtained by applying the registration-derived ground-truth displacement field. TRE was then computed as the Euclidean distance between $\hat{P}^i_{\text{intra}}$ and $P^i_{\text{intra}}$ in physical space (measured in millimetres). Together, Dice and TRE provide a comprehensive evaluation of global overlap accuracy and local geometric precision, respectively.

\section{Results}
\begin{figure}
\centering
\includegraphics[width=8cm]{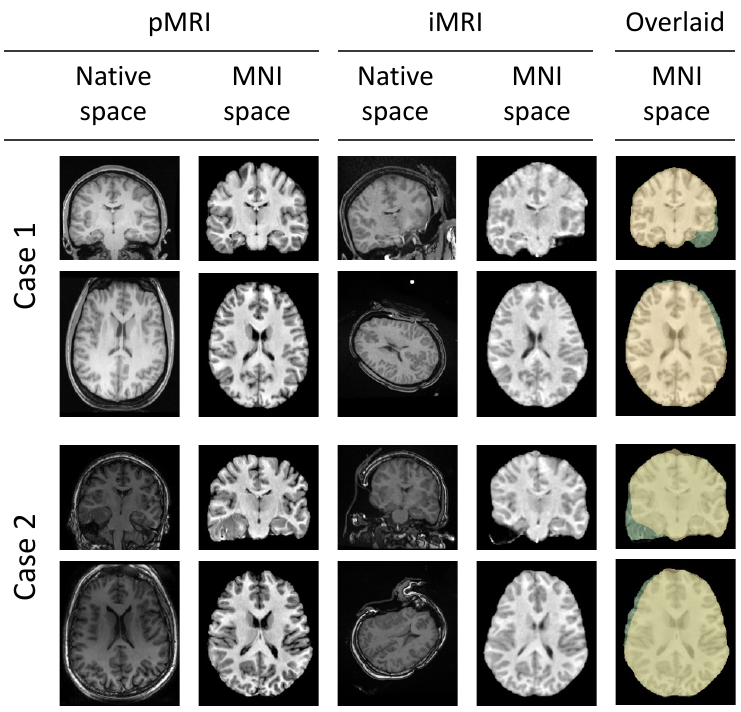}
\caption{Two representative examples of preoperative and intraoperative MRI normalisation into MNI space. For each case, pMRI and iMRI are shown in native space and after transformation to MNI space. The rightmost column overlays pMRI and iMRI in MNI space, demonstrating improved spatial correspondence after applying the preprocessing pipeline (Fig.~1) while preserving the post-resection cavity in iMRI.}

\label{fig:pipeline_results}
\centering
\end{figure}
\subsection{Pre-to-Intraoperative Registration Pipeline}
Fig.~\ref{fig:pipeline_results} provides qualitative examples of the preprocessing pipeline. For two representative cases, pMRI and iMRI are shown in native space and after transformation to MNI space. The MNI-space overlay illustrates improved spatial correspondence between modalities while retaining the post-resection cavity in iMRI, supporting the suitability of the pipeline as a preprocessing step for learning-based deformation prediction.

\subsection{Model performance}
\begin{figure}
\centering
\includegraphics[width=1.0\linewidth]{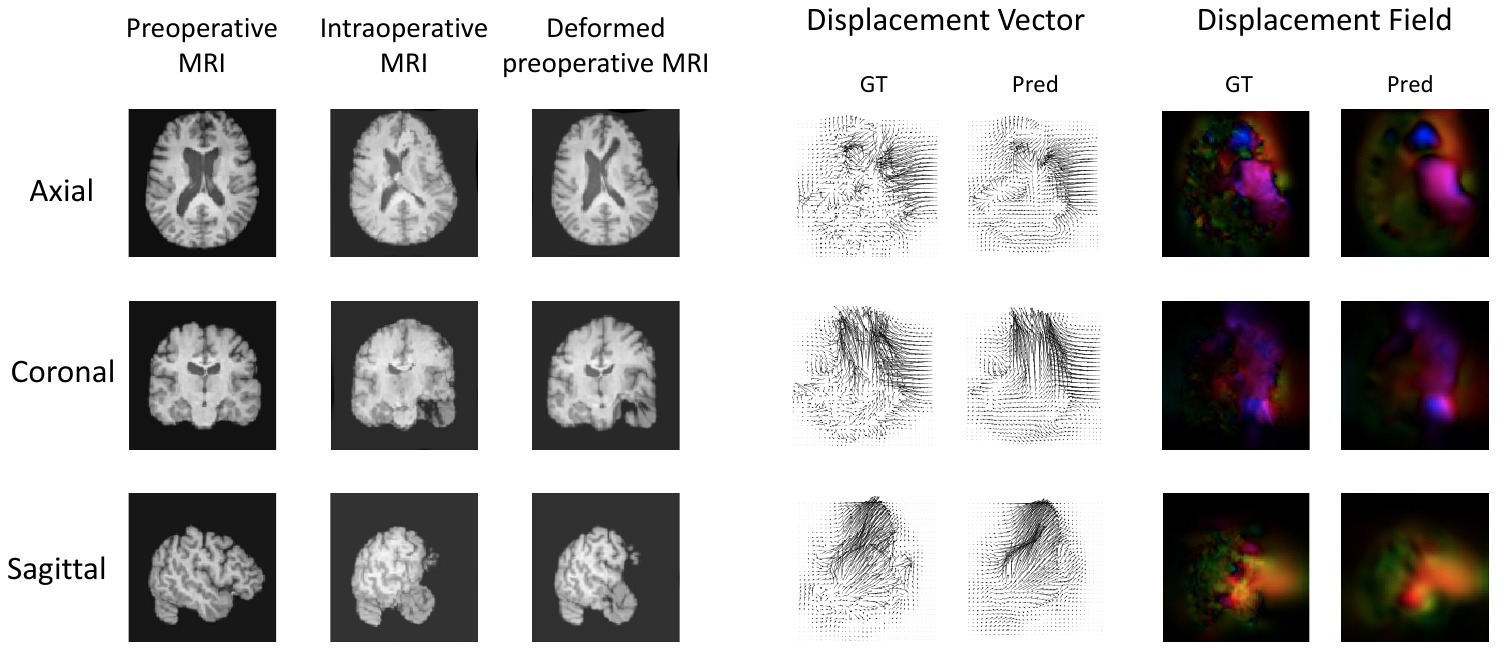}
\caption{Qualitative visualisation of predicted brain deformation in axial, coronal, and sagittal planes. The first three columns show the preoperative MRI, the intraoperative MRI, and the deformed preoperative MRI obtained using the predicted displacement field. The remaining columns compare registration-derived and predicted displacement fields, visualised using displacement vectors and optical flow.}

\label{fig:visualisation}
\centering
\end{figure}

We evaluated the performance of the proposed model from three complementary perspectives: (1) qualitative agreement between the deformed preoperative MRI obtained using the predicted displacement field and the intraoperative MRI; (2) qualitative consistency between the predicted and registration-derived displacement fields; and (3) accuracy of the predicted intraoperative brain mask and SDF image.

As shown in Fig.~\ref{fig:visualisation}, the deformed preoperative MRI obtained
using the predicted displacement field exhibits visually improved alignment
with the intraoperative MRI compared to the original preoperative image.
Qualitative agreement can be observed in multiple anatomical regions,
including inward displacement of the cortical surface in the axial view and
the appearance of the temporal lobe resection cavity in the axial and sagittal
planes.

The predicted displacement field is visualised using optical flow and
displacement vectors to illustrate both global deformation patterns and local
directional consistency. When compared with the registration-derived
displacement field used for supervision, the predicted field shows similar
spatial trends and deformation directions across axial, coronal, and sagittal
views, indicating consistent learning of the underlying deformation patterns.

\subsection{Evaluation of Brain Shift Compensation}
\begin{table}[t]
\centering
\small
\renewcommand\arraystretch{1.15}
\setlength{\tabcolsep}{3.5pt}
 \resizebox{\textwidth}{!}{%
\begin{tabular}{llcccccc}
\toprule
Resection & Comparison & P1 & P2 & P3 & P4 & P5 & P6 \\
\midrule
\multirow{2}{*}{Left}
& Preop $\rightarrow$ Intra 
& $4.58 \pm 1.47$ 
& $4.31 \pm 1.39$ 
& $1.62 \pm 0.55$ 
& $4.76 \pm 1.53$ 
& $1.48 \pm 0.50$ 
& $4.49 \pm 1.46$ \\
& Pred $\rightarrow$ Intra  
& $\mathbf{2.96 \pm 1.08}$ 
& $\mathbf{2.78 \pm 1.01}$ 
& $\mathbf{1.21 \pm 0.42}$ 
& $\mathbf{3.05 \pm 1.09}$ 
& $\mathbf{1.15 \pm 0.38}$ 
& $\mathbf{2.87 \pm 1.02}$ \\
\midrule
\multirow{2}{*}{Right}
& Preop $\rightarrow$ Intra 
& $4.41 \pm 1.42$ 
& $4.18 \pm 1.35$ 
& $1.58 \pm 0.53$ 
& $4.63 \pm 1.49$ 
& $1.46 \pm 0.48$ 
& $4.32 \pm 1.41$ \\
& Pred $\rightarrow$ Intra  
& $\mathbf{2.89 \pm 1.03}$ 
& $\mathbf{2.71 \pm 0.97}$ 
& $\mathbf{1.18 \pm 0.40}$ 
& $\mathbf{2.98 \pm 1.04}$ 
& $\mathbf{1.12 \pm 0.36}$ 
& $\mathbf{2.79 \pm 0.96}$ \\
\bottomrule
\end{tabular}
}
\caption{Mean Target Registration Error (TRE, in millimetres) under 9-fold cross-validation for landmarks located on the resection side and along the midline (P3 and P5). Results are grouped by the laterality of temporal lobe resection. For laterally paired landmarks (P1, P2, P4, and P6), TRE is reported for the ipsilateral hemisphere. Values are presented as mean $\pm$ standard deviation across folds.}

\label{tab:results_TRE}
\end{table}

\begin{table}[t]
\centering
\renewcommand\arraystretch{1.15}
\begin{tabular}{lcccccccccc}
\toprule
Mask Comparison & F1 & F2 & F3 & F4 & F5 & F6 & F7 & F8 & F9 & Avg \\
\midrule
$M_{\text{preop}}$ vs.\ $M_{\text{intra}}$ 
& $0.93$ & $0.94$ & $0.92$ & $0.93$ & $0.92$ 
& $0.93$ & $0.92$ & $0.93$ & $0.91$ 
& $0.92 \pm 0.01$ \\
$\hat{M}_{\text{intra}}$ vs.\ $M_{\text{intra}}$ 
& $\mathbf{0.98}$ & $\mathbf{0.97}$ & $\mathbf{0.96}$ & $\mathbf{0.97}$ & $\mathbf{0.96}$ 
& $\mathbf{0.97}$ & $\mathbf{0.96}$ & $\mathbf{0.97}$ & $\mathbf{0.95}$ 
& $\mathbf{0.97 \pm 0.01}$ \\
\bottomrule
\end{tabular}
\caption{Dice similarity coefficients for intraoperative brain mask prediction under 9-fold cross-validation. Dice scores are reported for each fold (F1--F9), with values summarised as mean $\pm$ standard deviation across folds.}

\label{tab:results_Dice}
\end{table}

Quantitative evaluation of brain shift compensation was performed using
landmark-based Target Registration Error (TRE) and Dice similarity coefficient,
as defined in Section~2.4. For each preoperative landmark $P_{\text{preop}}$, the predicted intraoperative
position $\hat{P}_{\text{intra}}$ was obtained by applying the predicted
displacement field. TRE was computed as the Euclidean distance between
$\hat{P}_{\text{intra}}$ and the corresponding intraoperative landmark position
obtained using the registration-derived displacement field, measured in
physical space (millimetres).

As reported in Table~\ref{tab:results_TRE}, landmark-based TRE is evaluated on
anatomical points located on the resection side together with midline landmarks
(P3 and P5). This design reflects the clinical setting of temporal lobe
resection for epilepsy surgery, where larger deformations are typically
observed near the surgical corridor, while contralateral displacement tends to
be comparatively small. In addition, neurosurgical decision-making primarily
depends on accurate guidance in the vicinity of the resection region and
adjacent critical structures.

Under 9-fold cross-validation, applying the predicted displacement field
consistently reduces TRE compared to the undeformed preoperative configuration,
indicating improved local geometric alignment in clinically relevant regions.

In addition to landmark-based evaluation, global brain shape agreement was
assessed using the Dice similarity coefficient. Table~\ref{tab:results_Dice}
summarises the Dice scores between the predicted intraoperative brain mask and
the ground-truth intraoperative mask, as well as between the preoperative and
intraoperative masks. Higher Dice scores are observed for the predicted masks
under 9-fold cross-validation, indicating improved overlap with the
intraoperative brain contour.

\section{Discussion}
During temporal lobe resection, brain shift is known to occur predominantly at
specific stages of the surgical procedure, particularly following dural
opening and upon ventricular exposure. As a result, the deformation observed
in intraoperative MRI (iMRI) reflects brain configuration at a later surgical
stage and can serve as a proxy for deformation trends relevant to the vicinity
of critical anatomical structures, such as the optic radiation. While the
timing of iMRI acquisition varies across procedures, the observed deformation
patterns provide valuable information for modelling brain shift in the context
of temporal lobe surgery.

A key aspect of the proposed approach is the joint prediction of a dense
displacement field and an intraoperative brain mask. This combined
representation enables the model to capture both voxel-wise deformation
patterns and global brain shape changes. By integrating these complementary
supervisory signals within a unified learning framework, the proposed method is
able to learn deformation patterns that are consistent at both local and global
scales.

It is important to distinguish the proposed approach from classical deformable
registration methods. Conventional registration algorithms require paired
images, typically a preoperative and an intraoperative scan, and aim to estimate
a transformation that aligns the two images under a chosen similarity criterion.
In contrast, the proposed method addresses a fundamentally different problem:
predicting plausible intraoperative deformation patterns using only preoperative
information, without access to intraoperative images at inference time. As such,
our objective is not post-hoc image alignment, but preoperative prediction of
future brain deformation, which places the task outside the standard formulation
of deformable registration. Although registration-derived deformation fields are
used as surrogate supervision during training, the inference setting and problem
formulation are fundamentally different from those of image registration.

Several limitations of this study should be acknowledged. First, the dataset
size is relatively limited, which may affect the generalisability of the
results across diverse surgical cases. In addition, the cohort used for
training and evaluation may not fully capture the variability present in
different clinical scenarios. Furthermore, the current study focuses on
predicting brain shift using preoperative MRI information alone, primarily to
investigate the predictive capability of such inputs. Incorporating additional
intraoperative modalities, such as intraoperative ultrasound or
electrophysiological measurements, could potentially provide complementary
information and improve deformation estimation. Finally, the evaluation metrics employed in this work, including Target
Registration Error (TRE) and Dice similarity coefficient, quantify geometric
alignment and mask overlap but do not fully capture clinical usability. A more
comprehensive assessment would require prospective clinical validation, such
as surgeon-in-the-loop evaluation or analysis of surgical outcomes, to better
understand the practical impact of the proposed method.

\section{Conclusions and Future Work}

In this work, we presented a learning-based framework that integrates a
preprocessing pipeline and a U-Net architecture to predict brain deformation
from preoperative to intraoperative MRI in the context of temporal lobe
resection surgery. The proposed approach models brain shift through dense
displacement field prediction and demonstrates improved geometric alignment of
key anatomical structures under the evaluated experimental settings.

Future work will focus on extending the proposed framework to incorporate
additional intraoperative observations, such as intraoperative ultrasound, to
further enhance deformation modelling in data-limited scenarios. In addition,
future studies will aim to evaluate the clinical utility of the method through
broader validation on larger cohorts and task-specific surgical assessments.

\noindent \\
\bibliography{sn-bibliography}

\end{document}